\documentclass[]{spie}  
\usepackage[]{graphicx}

\usepackage{appendix}
\usepackage{xcolor}
\usepackage{textcomp}

\title{Unsupervised Feature Learning in Remote Sensing} 

\author{Aaron Reite\supit{a}\footnote{\hspace{2mm}These authors contributed equally to this work.}, Scott Kangas\supit{b}\footnotemark[1], Zackery Steck\supit{b}, \\ Steven Goley\supit{b}, Jonathan Von Stroh\supit{c}, and Steven Forsyth\supit{d}, 
\skiplinehalf
\supit{a}NGA Research, 7500 GEOINT Dr., Springfield, VA, USA;\\
\supit{b}Etegent Technologies, Ltd., 5050 Section Ave., Suite 110, Cincinnati, OH, USA;\\
\supit{c}CACI, 15955 E Centretech Pkwy., Aurora, CO, USA;\\
\supit{d}NVIDIA, 2788 San Tomas Expressway, Santa Clara, CA, USA\\
}

\authorinfo{Further author information: (Send correspondence to A.R. and S.K.)\\  
A.R.: E-mail: aaron.a.reite@nga.mil\\ 
S.K.: E-mail: scott.kangas@etegent.com\\
}

\begin{document} 
  \maketitle 

\begin{abstract}
The need for labeled data is among the most common and well-known practical obstacles to deploying deep learning algorithms to solve real-world problems. The current generation of learning algorithms requires a large volume of data labeled according to a static and pre-defined schema. Conversely, humans can quickly learn generalizations based on large quantities of unlabeled data, and turn these generalizations into classifications using spontaneous labels, often including labels not seen before. We apply a state-of-the-art unsupervised learning algorithm to the noisy and extremely imbalanced xView data set to train a feature extractor that adapts to several tasks: visual similarity search that performs well on both common and rare classes; identifying outliers within a labeled data set; and learning a natural class hierarchy automatically.
\end{abstract}



\section{Introduction}
\label{sec:intro}  
In recent years, the-state-of-the-art for nearly every benchmark task in computer vision has been accomplished with deep convolutional neural networks (CNNs) trained via supervised learning on large, labeled data sets.\cite{alexnet,imagenet} However, obtaining high quality labeled data sets at the scale required for successfully training deep CNNs is costly, or even impossible, for many real-world computer vision applications, such as those requiring proprietary data, expert human data labeling, or those with limited real-world examples. 

Additionally, most of these CNNs are trained in static scenarios where a fixed number of labels are encountered, and these labels are consistent between training and testing. In practice, this ``closed-world'' assumption is often violated: instances may contain multiple classes making hard labels problematic; the domain may change as new classes arise; more fine-grained labels may be desired, etc. Transfer learning is a common method to address these issues; starting with a trained model, the final layers are removed and replaced with randomly initialized layers before undergoing additional training to adapt to the new data distribution. Of course, this new model requires even more labeled data and is subject to its own ``closed-world'' so is similarly brittle if the test distribution changes again.

\begin{figure}[htbp]
\centering
\includegraphics[scale=1.7]{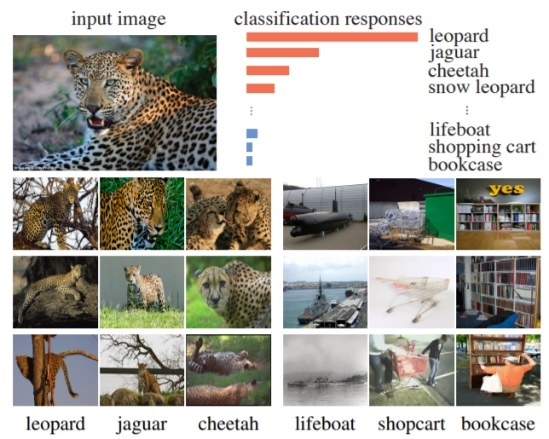}
\caption{Results from an ImageNet CNN trained with supervision: inference on a leopard class instance returns high responses from visually similar classes (from Wu et al \cite{ufl}).}
\label{fig:leopard}
\end{figure}

In this paper, we investigate unsupervised learning techniques applied to remotely sensed imagery, with the dual  goal  of  training  a  feature  extractor  that  is  readily adaptable to new tasks  and that does  not  depend  on  any labeled  data  for  training.   Specifically, we use Unsupervised Feature Learning via Non-Parametric Instance-level Discrimination (UFL) developed by Wu, Xiong, Yu and Lin at the University of California Berkeley, which surpassed the state-of-the-art unsupervised learning methods for classification on ImageNet 1K in 2018\cite{ufl}. The authors of UFL observed that classes that appear visually correlated generally receive higher softmax output scores than classes that are visually uncorrelated (e.g., Figure \ref{fig:leopard}).  Based on this observation, the authors developed an unsupervised learning approach to discriminate between individual instances, completely ignoring class labels and allowing the network to learn similarity between instances without a need for semantic categories. Treating each instance as a class results in computational challenges: for ImageNet the number of ``classes'' expands from the original 1000 to 1.2 million, i.e. the number of images in the training set. The authors of UFL address these computational challenges with a low-dimensional memory bank and a noise-contrastive estimate of their non-parametric softmax classifier. We discuss the UFL method in detail in Section \ref{sec:ufl}.

The data we use in our experiments is derived from the xView object detection data set\cite{xview}. Although originally intended to advance the state-of-the-art for object detection in overhead imagery, we use the provided annotation bounding boxes to extract image chips around each object for the purpose of classification. Further detail about xView and our extracted image chips is covered in Section \ref{sec:data}.

We use our UFL trained feature extractor for several, diverse computer vision tasks. First, we emulate the UFL paper's unsupervised classification experiment on our data in \ref{sec:classification}. We find that UFL greatly outperforms a common competing unsupervised method (autoencoder), and even beats a fully supervised counterpart in top-5. We then demonstrate the generalizability of our UFL trained feature extractor by applying it to three disparate tasks: we show that a UFL trained feature extractor may be used in a similarity search algorithm that performs well even with highly imbalanced classes in \ref{sec:simsearch}; we identify outliers and errors in xView class labels in \ref{sec:outliers}; and we learn a visual hierarchy automatically in \ref{sec:hierarchies}.




\section{Approach} 
\label{sec:approach}
\subsection{Data Set}
\label{sec:data}

XView is the Defense Innovation Unit (DIU) and the National Geospatial-Intelligence Agency's (NGA) large-scale detection data set that contains 1,413 km\textsuperscript{2} of 3-band panchromatic sharpened WorldView-3 satellite imagery. The images were collected at 0.3m ground sample distance, with 60 types of objects and land use categories exhaustively labeled with axis aligned bounding boxes, for a total of 1M+ bounding boxes.\cite{xview} During the 2018 xView Detection Challenge, a significant portion of these data were publicly released: 847 1km\textsuperscript{2} images with 601,937 bounding box labels for training, and 281 1km\textsuperscript{2} additional images without labels\cite{xview_webpage}. We procured the bounding box labels for these 281 additional images for our research, which we reserved for testing.

\begin{figure}[htp]
\centering
\includegraphics[width=17cm]{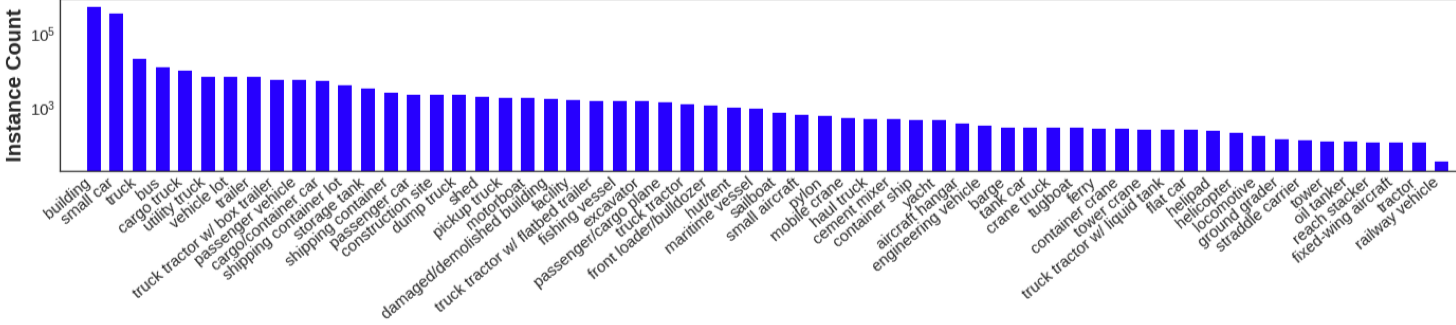}
\caption{Instances per class in xView (from Lam et al \cite{xview}).}
\label{fig:xview_class_count}
\end{figure}

The objects in xView present unique challenges that differ from standard detection tasks such as Microsoft's Common Objects in Context (COCO)\cite{COCO}. For example, most of the xView objects are very small (10s of pixels) and have no standard orientation---they are rotated in different directions. Perhaps most challenging is the extreme class imbalance present in xView. If the data were class balanced, each of the 60 classes would have 10K examples for training. However, just two classes (Small Car and Building) account for almost 90\% of the training data, so most classes have far fewer examples: many have 100s of examples, while some have just 10s. Figure \ref{fig:xview_class_count} demonstrates xView's extreme class imbalance. The first place solution to the 2018 xView Detection Challenge cites class imbalance as the most difficult obstacle to building a successful detector, and develops a novel loss function based on Focal Loss to assist\cite{xview-1stplace, FocalLoss}.  

We created a classification data set from xView by cropping square image chips centered around each bounding box, using the box's longest dimension. We chose square chips to allow ingestion into standard CNNs without resizing each dimension independently, and to allow some neighboring pixels for each object in the chip to provide context. For example, boat chips include neighboring water pixels and car chips include neighboring road pixels. We explored alternative methods to create image chips that would allow varying amounts of context and still provide square chips, such as taking rectangular crops proportionately longer than each side of the bounding box and then zero-padding to make a square; however, we found no measurable benefit in our experiments to these more complicated methods. We discarded chips that cross image boundaries and therefore contain incomplete objects. Table~\ref{tab:xview} contains a few key statistics about our data set.

Like many computer vision data sets, xView contains labeling errors. While deep learning has proven to be robust to labeling errors in well-sampled classes, the effect of these errors increases dramatically in classes containing few instances. As an extreme example, our test set for Railway Vehicle contains just two instances, both of which are labeled incorrectly (they are identifiable as specific types of railway vehicles). Nonetheless, after extensive experimentation with customized classes derived from the original 60, we eventually elected to keep all 60 classes as designated by the original xView paper to ease repeatability. Accordingly, our results may be significantly improved by eliminating a few very small or erroneous classes (e.g., Railway Vehicle), merging a few classes that are visually indistinct (e.g., Small Car and Passenger Vehicle), and splitting a few classes that contain visually distinct sub-classes (e.g., Tower).    

\begin{table}[h]
\caption{Summary statistics for our classification data set derived from xView.} 
\label{tab:xview}
\begin{center}       
\begin{tabular}{|l|l|l|} 
\hline
\rule[-1ex]{0pt}{3.5ex}   & \textbf{Training} & \textbf{Test} \\
\hline
\rule[-1ex]{0pt}{3.5ex}  Minimum Chips per Class & 17 & 2  \\
\hline
\rule[-1ex]{0pt}{3.5ex}  Median Chips per Class & 629 & 125  \\
\hline
\rule[-1ex]{0pt}{3.5ex}  Maximum Chips per Class & 307221 & 100899  \\
\hline
\rule[-1ex]{0pt}{3.5ex}  Total Chips & 589119 & 187156  \\
\hline
\end{tabular}
\end{center}
\end{table}
\subsection{Unsupervised Feature Learning (UFL)}\label{sec:ufl}

\begin{figure}[htbp]
\centering
\includegraphics[width=15cm]{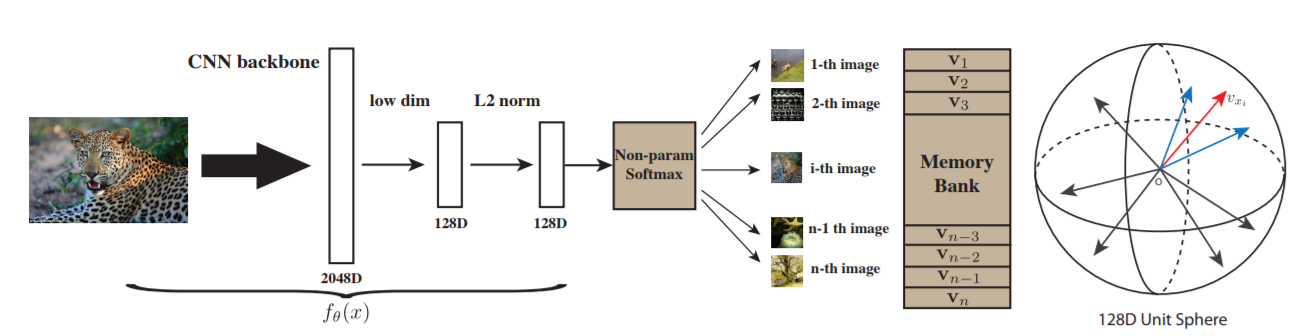}
\caption{Pipeline for the Unsupervised Feature Learning approach (from Wu et al \cite{ufl}).}
\label{fig:ufl_pipeline}
\end{figure}

Figure \ref{fig:ufl_pipeline} shows the overall pipeline for the UFL approach. A standard CNN is used to embed each image as a feature vector, which is then $L^2$ normalized prior to being passed to a non-parametric softmax classifier for instance level discrimination (described in Section \ref{nce}).  The feature embedding is trained to maximally distribute the embedded features over the unit hypersphere.  For evaluation, the authors implemented a KNN classifier using cosine similarity between the feature vector for a test image and those for the images used during training.

Benefits of this unsupervised approach are two-fold: annotations are not needed at training time thereby eliminating the burdensome task of data labelling to train a feature extractor, and the method is agnostic to network architecture so it can be implemented on any current or future state-of-the-art network design.

\subsubsection{Non-Parametric Instance Discrimination}\label{nce}
Let $f_\theta$ be a CNN with parameters $\theta$ mapping images $x_i$ to feature vectors $v_i = f_\theta(x_i)$. A conventional parametric softmax classifier will classify image $x$ as class $i$ with probability:
\begin{equation}\label{eqn:softmax}
    P(i|v) = \frac{e^{w_i^T v}}{\sum\limits_{j=1}^n e^{w_j^T v}}
\end{equation} where $n$ is the number of classes, $w_j$ is a learned weight vector for class $j$, and the inner product $w_j^Tv$ computes the similarity between $v = f_\theta(x)$ and class $j$.

Instead of computing the similarity between a feature vector $v$ and the class weight vectors $w_j$, which may be interpreted as class prototypes, the authors of UFL modified Equation \ref{eqn:softmax} to compute the similarity of $v$ to other feature vectors: $v_j^Tv$. As all feature vectors have unit norm, this is the cosine similarity between $v$ and $v_j$. Thus, in UFL Equation \ref{eqn:softmax} becomes:

\begin{equation}\label{eqn:nonparam_softmax}
    P(i|v) = \frac{e^{(v_i^T v)/\tau}}{\sum\limits_{j=1}^n e^{(v_j^T v)/\tau}}
\end{equation} where $\tau$ is a parameter controlling the concentration of the $v_j$ on the unit sphere and $n$ is the total number of instances in the training set. $P(i|v)$ approaches $1$ if $v$ is close to $v_i$ and far away from all other $v_j$, or approaches $0$ if $v$ is far away from $v_i$ (so long as $\tau$ is chosen to be sufficiently small: we used the UFL paper's recommended value $\tau = 0.07$).

Unfortunately, calculating the non-parametric softmax in Equation \ref{eqn:nonparam_softmax} may be computationally prohibitive when each image is considered its own class, e.g. ImageNet has 1.2M annotated images all of which require feature vectors for the computation. Instead of repeatedly computing each feature vector $v_j$, the feature vectors are stored in a memory bank $V = \{v_j \mid 0 \leq j \leq n\}$. The dimension of the feature space is chosen to be relatively low: the hypersphere in ${\rm I\!R^{128}}$ (in both our work and the UFL paper, differing values for the feature dimension were explored, but none offered significant benefit). This choice allows all 1.2M feature vectors in ImageNet to require only 600 Mb of memory.

The learning objective is to minimize the negative log-likelihood over the training set: \begin{equation}\label{eqn:loss}
    J(\theta) = -\sum_{i=1}^{n} \log P(i|f_\theta(x_i)). 
\end{equation} The loss therefore depends on how close each $f_\theta(x_i)$ is from $x_i$'s feature vector during the previous iteration, $v_i$, as well as how far $f_\theta(x_i)$ is from all the other feature vectors $v_j$. As such, a minimal solution for $\theta$ will result in an equidistribution of $v_j$ over the sphere. As $f_\theta$ has a finite number of convolutional filters and the embedding space $S^{127}$ is compact, a minimal solution forces the $v_j$ which activate the same convolutional filters (i.e., are visually similar) to be close together. During each learning iteration, all network parameters $\theta$ and the feature vector $f_\theta(x_i)$ are updated via stochastic gradient descent and $v_i \in V$ is replaced with $f_\theta(x_i)$.

The burden of computing Equations \ref{eqn:nonparam_softmax} and \ref{eqn:loss} may be even further, and dramatically, reduced by employing  noise-contrastive estimation to replace the denominator of Equation \ref{eqn:nonparam_softmax} with a constant computed from a Monte Carlo approximation during the initial few batches\cite{nce}. This reduces the learning objective to a much simpler form. We use the same normalizing constant and computational approximations as the UFL authors and  refer the reader to Section 3.2 of Wu et al for additional details\cite{ufl}.

Note that if random augmentation is employed during training, such as flips, rotations, color jitter, etc.,  then each feature vector $v_i \in V$ may be interpreted as a class prototype for the class created from image $x_i$ by applying random augmentations. As such, $f_\theta(x_i)$ can not equal $v_i$ even if the network parameters $\theta$ remain unchanged. This provides a consistent learning signal to teach the network to become invariant  to augmentations, rather than just spread the feature vectors evenly across the sphere.   

\subsubsection{Weighted KNN Classification}\label{sec:knn}
In order to classify an image $\hat{x}$ in our validation set, we first compute its feature $\hat{v} = f_{\theta}(\hat{x})$ and compare it to all of the feature vectors $v_i \in V$ using cosine similarity: $v_i^T\hat{v}$. We then take the $k$ feature vectors in the memory bank that are closest to $\hat{v}$ with respect to cosine similarity---the $k$ nearest neighbors, $\mathcal{N}_k$. Finally, $\hat{x}$ is classified by a weighted voting of the classes in $\mathcal{N}_k$; specifically, if we denote the class of image $x_j$ corresponding to memory bank vector $v_j \in \mathcal{N}_k$ by $c_j$, then class $c_j$'s vote is: 

\begin{equation}
\sum\limits_{v_i \in \mathcal{N}_k}\delta_{c_i,c_j} e^{(v_i^T\hat{v})/\tau} \textrm{, where } \delta_{c_i,c_j}=
\left\{
   \begin{array}{ll}
   1 & {\rm when~} c_i=c_j\\
   0 & {\rm when~} c_i\ne c_j
   \end{array}
\right\}
\end{equation} where $\tau$ controls how $v_i$'s distance from $\hat{v}$ effects its vote. With $\tau = 0.07$, as per training, a feature vector that is close to $\hat{v}$ will count $\sim$6x more than a feature vector that is 30 degrees away.  Nonetheless, $k$ must be chosen to limit the size of $\mathcal{N}_k$, particularly in the case of imbalanced classes; otherwise, the vote will be dominated by the most populous classes despite small values of $\tau$. We use $k=50$.

\section{Experiments}
 \label{sec:experiments}
 
In all of our experiments, we first learn a robust feature extractor in an unsupervised setting by applying UFL to our data set. The performance of this feature extractor is judged by the top-5 unsupervised classification score on our test set. 

\subsection{Unsupervised Classification}\label{sec:classification}

We use ResNet18 as the backbone CNN with a low dimension of 128. As noted, we use $\tau=0.07$ and $k=50$, with an initial learning rate of $0.03$. We quickly discovered that pre-training on ImageNet results in dramatic improvements; in fact, \textit{pre-training on ImageNet without fine-tuning readily beats training for 200 epochs from randomly initialized weights}, clearly demonstrating UFL's robustness to domain adaptation. Fine-tuning requires a smaller learning rate and an aggressive decay schedule (we use a learning rate of $0.001$ and decay by a factor $0.5$ every two epochs). In addition to our UFL models, we train two additional ResNet18-based models in order to establish baseline accuracy performance:

\begin{itemize}
    \item An autencoder with reconstruction loss and a 128-dimensional embedding space.  During evaluation the decoder stage of the network was removed and the weighted classification described in Section \ref{sec:knn} was implemented.
    \item A supervised model using softmax and cross-entropy loss and class-balanced sampling. This model represents the standard technique given the benefit of fully labeled data and should represent a ceiling on unsupervised performance.
\end{itemize} All models were trained with random initialization as well as fine-tuning after being pre-trained on ImageNet. 

Class-balanced sampling at train time as we used in our supervised model significantly improves our UFL results; however, we chose not to use this technique as it requires prior knowledge of the class labels and is contrary to the assumptions of unsupervised learning. 

Given the large disparity in class populations in our data set, we report our top-1 and top-5 results averaged over the 60 classes, instead of averaged over all instances. This is necessary because, as noted, the Small Car and Building classes alone count for 88\% of our data (36\% and 52\% respectively). Given a building in the test set, a random guess using the distribution of our training data will result in a correct classification 52\% of the time and a correct classification in 5 guesses (i.e., top-5) 97\% of the time. As such, a random network is expected to produce top-1 and top-5 scores of 40.1 and 83.2 (respectively) when averaged over all images, but only 1.7 and 4.1 (respectively) when averaged over all 60 classes. Table~\ref{tab:classification} summarizes the results from our classification models on our data set.

\begin{table}[h]
\caption{Classification using supervised and unsupervised learning methods with a ResNet18 backbone. Winning unsupervised method in bold.} 
\label{tab:classification}
\begin{center}       
\begin{tabular}{|l|l|l|} 
\hline
\rule[-1ex]{0pt}{3.5ex}  \textbf{Method} & \textbf{Top-1 Accuracy} & \textbf{Top-5 Accuracy}  \\
\hline
\rule[-1ex]{0pt}{3.5ex}  Supervised Random Init. & 35.8 & 50.5 \\
\hline
\rule[-1ex]{0pt}{3.5ex}  Supervised Fine-tuned & 42.4 & 65.6 \\
\hline
\rule[-1ex]{0pt}{3.5ex}  Autoencoder  Random Init. & 3.8  & 20.5 \\
\hline
\rule[-1ex]{0pt}{3.5ex}  Autoencoder  Fine-tuned. & 3.6  & 19.9 \\
\hline
\rule[-1ex]{0pt}{3.5ex}  UFL Random Init. &  6.8 & 28.9 \\
\hline
\rule[-1ex]{0pt}{3.5ex}  UFL Pre-trained, not Fine-tuned & 12.9 &  47.6 \\
\hline
\rule[-1ex]{0pt}{3.5ex}  \textbf{UFL Fine-tuned} & \textbf{18.3} & \textbf{54.5} \\
\hline 
\end{tabular}
\end{center}
\end{table}

All of the UFL models beat the autoencoders by significant margins. Notably, pre-training UFL on ImageNet \textit{without fine-tuning on xView} (UFL Pre-trained, not Fine-tuned) more than doubles the autoencoders' scores and produces a top-5 score comparable to our randomly initialized fully-supervised model. If fine-tuned, UFL beats our randomly initialized fully-supervised model's top-5 score by 8\%, \textit{despite having no labeled data during training}.

To examine the affects of increasing the backbone CNN's depth and capacity, we trained UFL using ResNet50 which resulted in higher scores: top-1 = 20.2 and top-5 = 55.2 (fine-tuned from ImageNet). We hypothesize that even deeper CNN backbones will result in better accuracy. However, the purpose of our work is to demonstrate generalizibility of UFL when applied to diverse tasks in remote sensing imagery, not set or best benchmark scores, so we use ResNet18 in our following experiments as it is nearly as accurate, but much faster with many fewer parameters.  Detailed class-by-class results for our fine-tuned ResNet18 UFL model are included in Table \ref{tab:classificationDetail} within Appendix \ref{sec:appendixClassificationResults}.

\subsection{Similarity Search}\label{sec:simsearch}

The challenge of similarity search, or image retrieval, is this: given a query image as input, return the image in a data set most similar to the query. Traditionally, Content-Based Image Retrieval (CBIR) has required large labelled data sets for training and expensive feature-point calculations to extract important visual information\cite{CBIR}. However, this expectation is becoming infeasible in a data-driven world where data sets may be petabytes in size; in such cases unsupervised deep learning approaches are preferable. The appeal of using UFL for image retrieval is that it does not require knowledge of class labels for training or image retrieval, thus it is completely unsupervised. Likewise, the foundation of UFL is that visually similar objects will be close together in the embedding space, making it an ideal candidate for CBIR.

We implement an image-retrieval algorithm using the fine-tuned UFL and autoencoder networks described in Section \ref{sec:classification} by modifying the weighted KNN classifier described in Section \ref{sec:knn}. Instead of collecting the nearest $k$ class labels for a test image (i.e. query), we collect the nearest $k$ instance indices, which are used to retrieve their respective images from the training data (i.e. query results). For both networks, we return the nearest five images for four query classes---Building, Small Car, Helicopter, and Aircraft. Helicopter and Aircraft are considered rare or low-shot classes, containing only 68 and 73 instances per class respectively.

\begin{figure}[htp]
\centering
\includegraphics[scale=0.25]{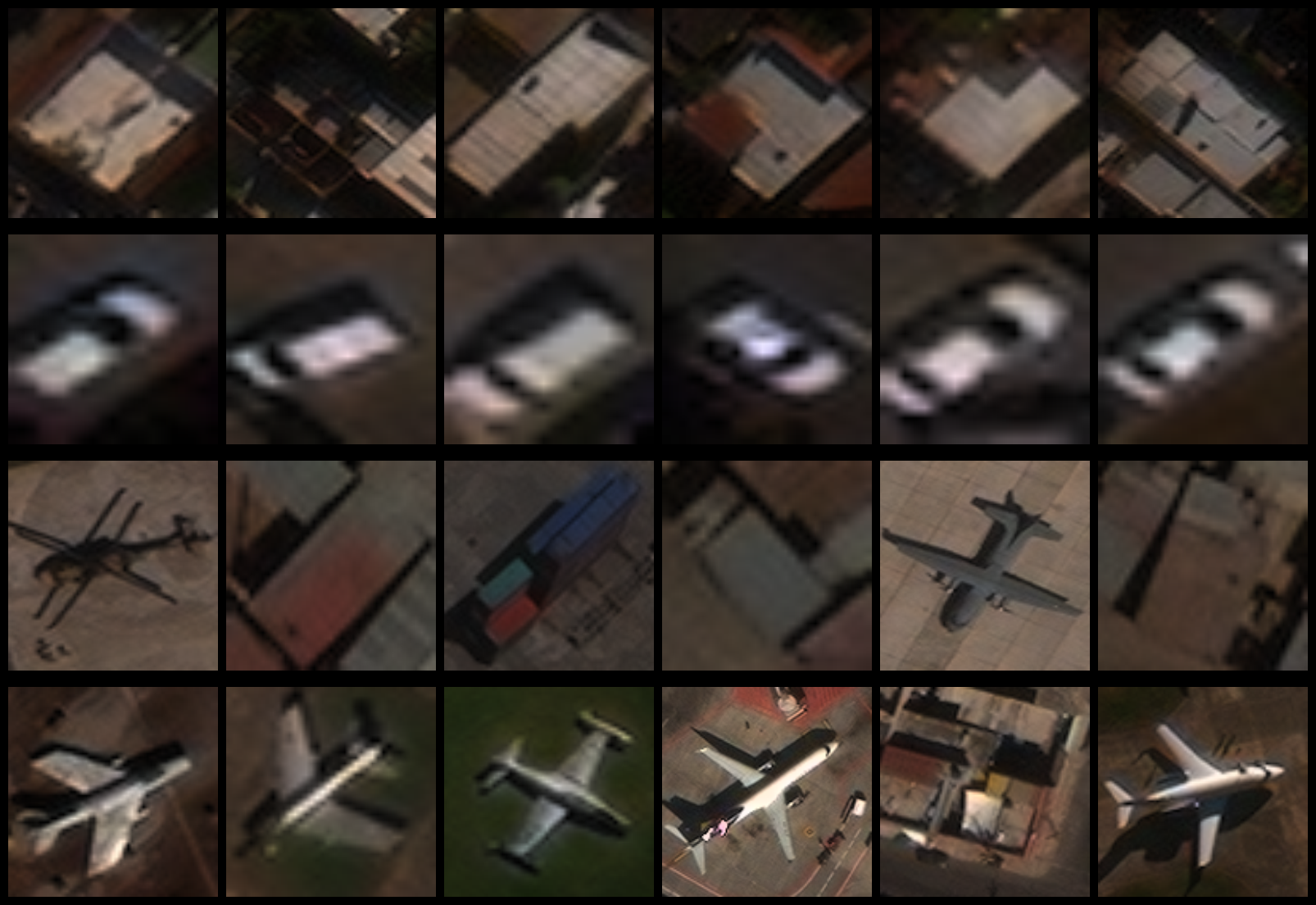}
\caption{Images retrieved by the UFL network. The query images are in the first column while the following five columns contain the closest five images in the training set.}
\label{fig:UFL_retreeval}
\end{figure}

Figure \ref{fig:UFL_retreeval} shows the UFL architecture performs well for the Building (row 1), Small Car (row 2), and Aircraft queries (row 4). Although the results for the Helicopter query (row 3) are incorrect, they all contain multiple visual similarities: dark orthogonal lines or shadows similar to rotors and small enclosed objects similar to a helicopter's fuselage. It's worth noting that the Aircraft query achieved acceptable results with only five more samples in its source class than the Helicopter class.

\begin{figure}[htp]
\centering
\includegraphics[scale=0.25]{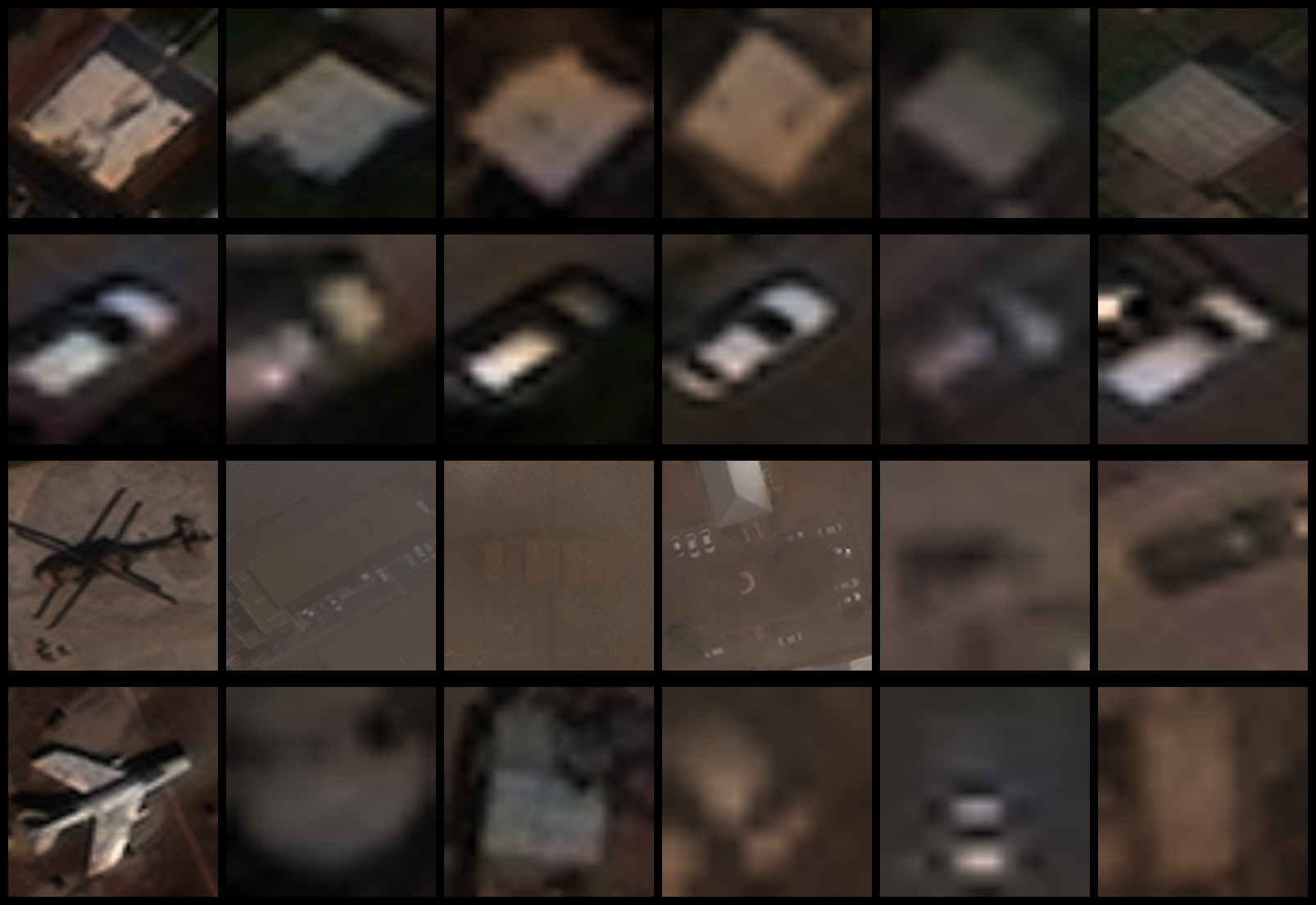}
\caption{Images retrieved by the autoencoder network. The query images are in the first column while the following five columns contain the closest five images in the training set.}
\label{fig:autoencoder_retrieval}
\end{figure}

Figure \ref{fig:autoencoder_retrieval} shows that the autoencoder obtains similar results for both Building and Small Car queries, but is outperformed by UFL in the rare or low-shot classes. The Helicopter query (row 3) and Aircraft query (row 4) lack any visual similarity to the query image beyond background color---reconstruction loss fails to embed visually similar objects in close proximity. 

\begin{figure}[htp]
\centering
\includegraphics[scale=0.40]{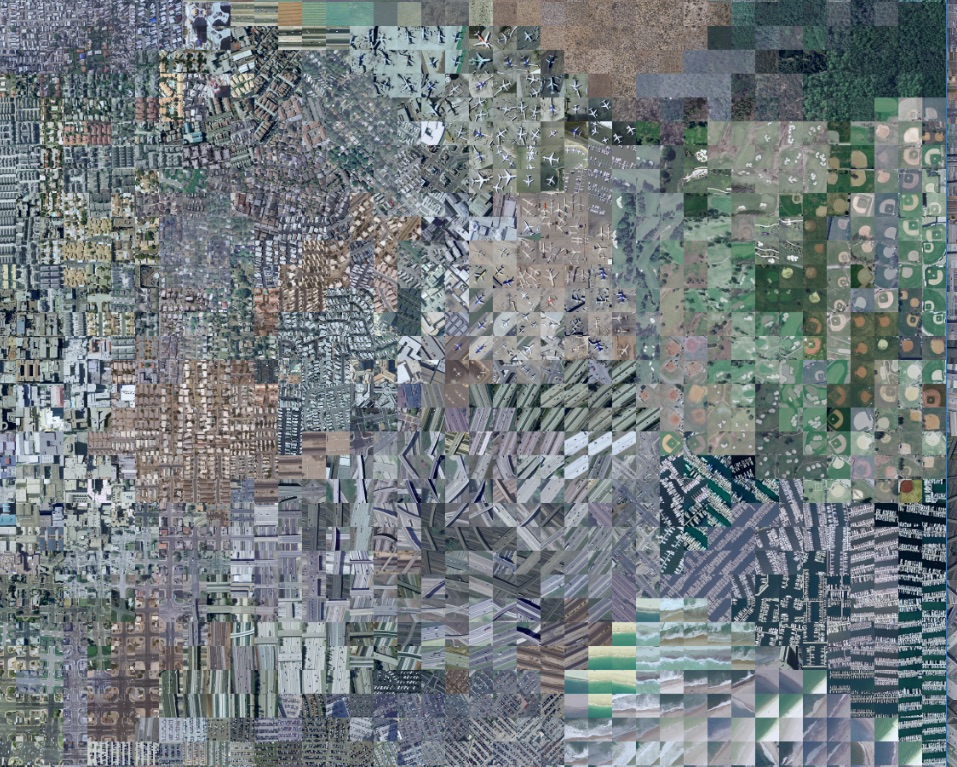}
\caption{UC Merced Land Use Dataset trained and embedded with UFL and visualized with a t-SNE grid.}
\label{fig:UFL-tsne}
\end{figure}

In addition to the similarity search using the xView dataset, we train UFL on the UC Merced Land Use Dataset and project the resultant feature vectors into a 2-D map, using t-SNE \cite{UCMerced, UCMerced_webpage,tsne}. As shown in Figure \ref{fig:UFL-tsne}, images with similar content have feature vectors that are close in the embedding space.

\subsection{Identifying Outliers}\label{sec:outliers} 
 
Procuring large annotated data sets often results in noisy labels, as we discussed for xView in \ref{sec:data}.  We use our UFL trained network to assist in identifying potentially erroneous labels.  As UFL does not make use of labels during training, it will not attempt to tighten intra-class samples nor will it repel inter-class samples; as such, it is a great candidate to detect outliers from a training set.  For this purpose, we (1) train UFL in the standard manner (fine-tuned from ImageNet), (2) create a $K-D$ tree for the training feature vector set, (3) compute the distance to nearest neighbor intra-class feature vectors for each instance, (4) compute the intra-class mean of all such nearest neighbor distances, and (5) identify instances with a distance greater than two standard deviations from their intra-class mean. 

\begin{figure}[!htp]
\centering
\includegraphics[scale=1.0]{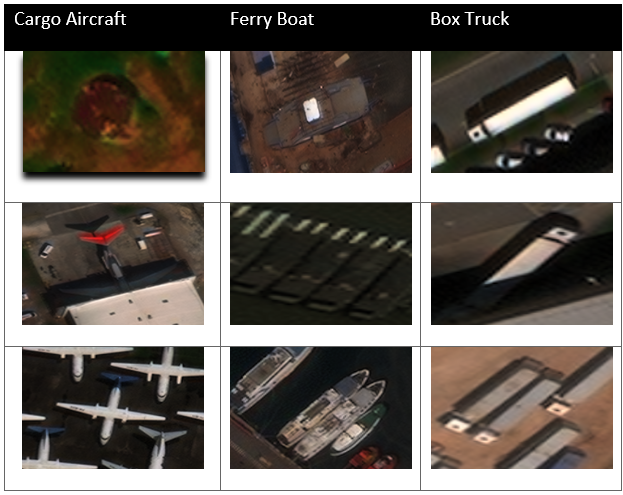}
\caption{Examples of class outliers identified using UFL (in no particular order).}
\label{fig:outliers}
\end{figure}

Figure \ref{fig:outliers} shows some examples of outliers found using this methodology. This technique is able to identify incorrectly labeled instances as well as anomalous but correctly labeled instances, such as those that are obscured, crowded or appear with unusual backgrounds.

\subsection{Unsupervised Learning of Visual Hierarchies}\label{sec:hierarchies}

A class hierarchy is a directed, acyclic graph where each class is a node, and edges between nodes express the relationship ``is a''; for example, a Locomotive is a Rail Vehicle. Many human-created hierarchies have been used for machine learning purposes, such as WordNet (from which ImageNet labels are derived), but these hierarchies do not always make sense from a computer vision perspective as they often use non-visual relationships\cite{wordnet}. For example, a human may create a hierarchy in which Trailer is close to Truck Tractor w/ Box Trailer, but far from Shipping Container. However, from the computer vision perspective, Trailer and Shipping Container are nearly identical---often they may only be distinguished if a trailer's wheels are visible, which may not be possible given the orientation of an imaging satellite. 


 We follow the method for learning hierarchies from a classifier's predictions developed by Silva-Palacios, Ferri, and Ram\'irez-Quintana at the Universitat Polit\`encia de Val\`encia\cite{hierarchies}. For brevity, we only describe the method at a high level: consult Section 2.2 of Silva-Palacios' paper for details. First, we create a confusion matrix $M$ from UFL classification predictions, as discussed in \ref{sec:classification}. We then create the similarity matrix $D$ from $M$ by applying three functions consecutively. $D$ is symmetric with entries between 0 and 1. Entries corresponding to classes that are commonly confused (or predicted correctly; i.e., the diagonal entries) are close to 0, while classes that are rarely confused have entries close to 1. Finally, we apply a standard agglomerative hierarchical clustering algorithm and plot the associated dendrogram.
 
 \begin{figure}[htp]
\centering
\includegraphics[scale=0.55]{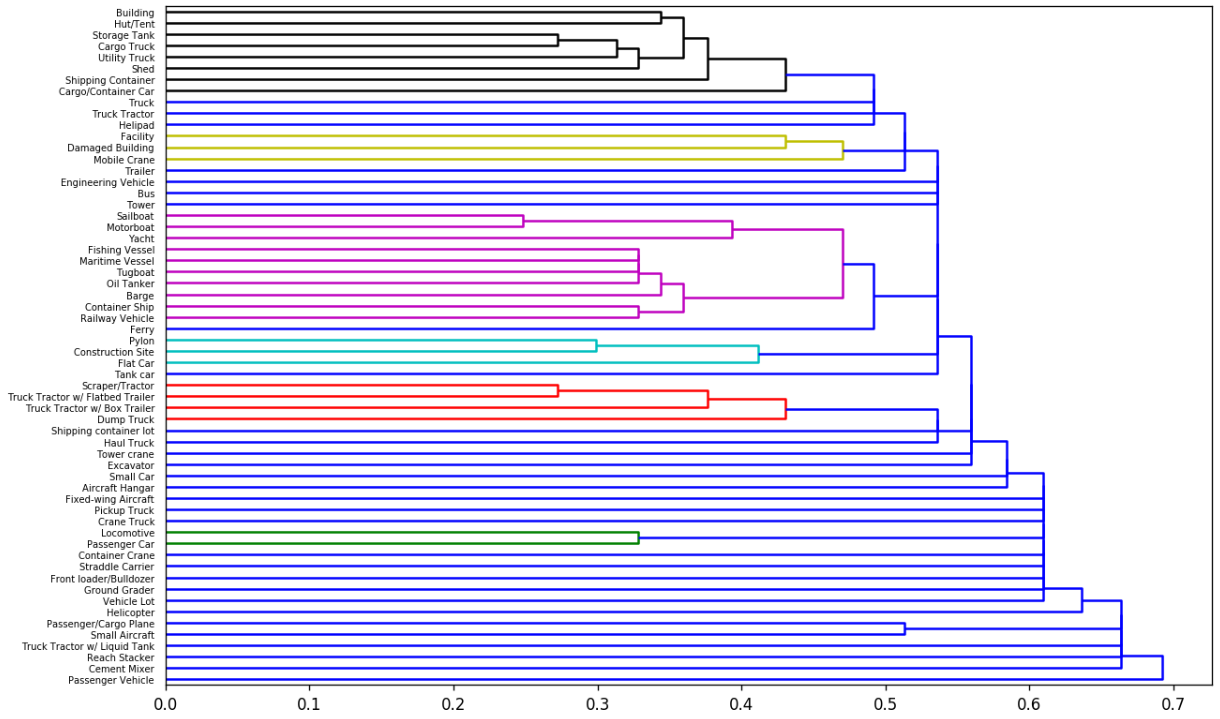}
\caption{Learned hierarchy using top-1 classification decisions.}
\label{fig:top1hierarchy}
\end{figure}

\begin{figure}[htp]
\centering
\includegraphics[scale=0.55]{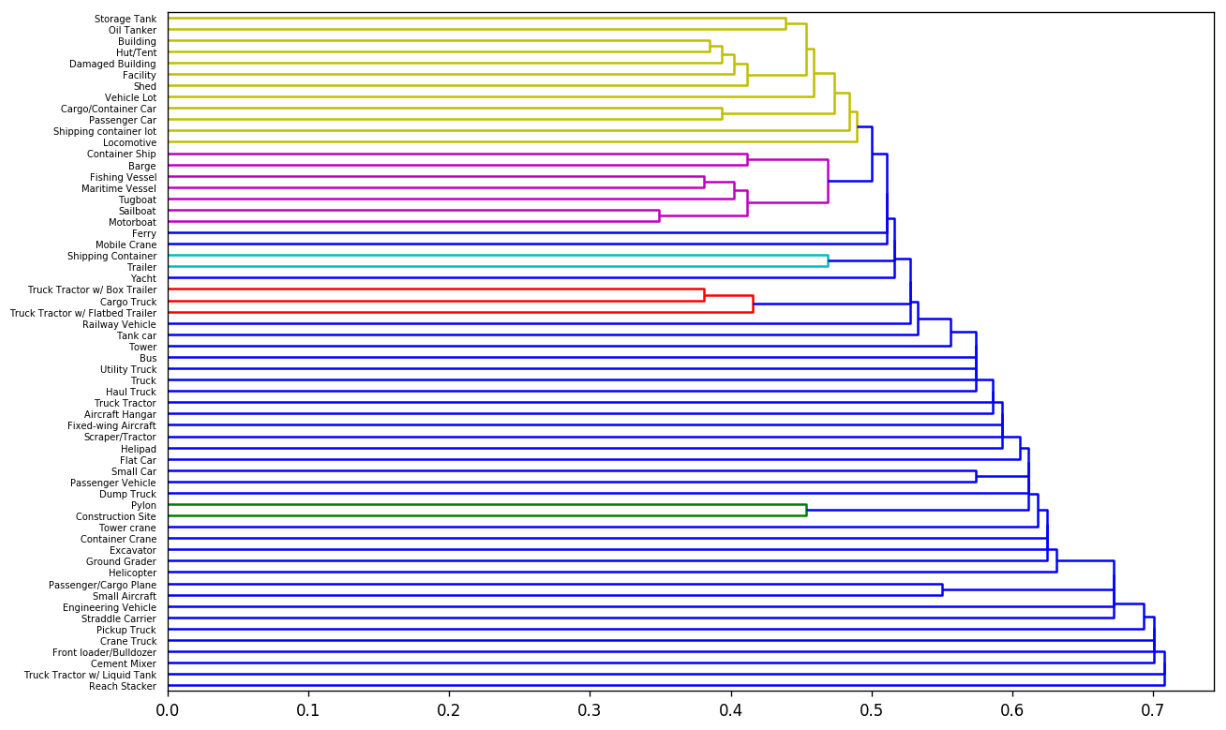}
\caption{Learned hierarchy using the top-5 classification decisions.}
\label{fig:top5hierarchy}
\end{figure}
 
 We perform this hierarchy learning for both top-1  and top-5 classification decisions (Figures \ref{fig:top1hierarchy} and \ref{fig:top5hierarchy}). We observe that some portions of the hierarchy are natural and intuitive: buildings seem to be closely related to other types of buildings, and the same for many types of vehicles, rail cars, and maritime vessels. Other relationships which are less intuitive also occur; for example, Helipad is closely related to both Truck Tractor and Truck in the top-1 hierarchy. One hypothesis is that most Helipads in xView are made of concrete with large, rectangular painted lines, while Truck Tractors and Trucks both are large, rectangular vehicles that are often surrounded by concrete.

\section{Summary}

In this paper, we compare UFL against similar unsupervised and supervised architectures for classification of objects in xView. We observe that UFL fine-tuned from ImageNet dramatically outperforms autoencoder models and approaches supervised model performance in top-5 accuracy. We also show that UFL trained models can be used for visual similarity searches that perform well on both common and rare or low-shot classes. Additionally, we use UFL to identify outliers within the labeled xView data set and to learn a visual class hierarchy automatically. 

In an upcoming paper, the authors will implement a hierarchical classifier using a Bayesian factor graph to estimate posterior probabilities over a spectrum of classes, ranging from very general at the top of the hierarchy to very specific at the bottom of the hierarchy. The goal is to improve accuracy given the benefit of a hierarchy, especially in cases of rare or low-shot classes. 

\acknowledgments     
 
This work was supported by NGA / NVIDIA Cooperative Research and Development Agreement\newline HM0476CRFY17007, NGA / Etegent Technologies Ltd. contract  HM047618C0071, and NRO / CACI contract 12-D-0227. It is approved for public release by National Geospatial Intelligence Agency \#19-863. We gratefully acknowledge the support of NVIDIA for donating a DGX-1 from its PSG cluster for this research. We also thank Kyle Pula and Jonathan Howe for many helpful discussions and comments.


\bibliography{main}   
\bibliographystyle{spiebib}   

\newpage
\appendix
\section{Unsupervised Classification Results}\label{sec:appendixClassificationResults}
\begin{table}[h]
\caption{Detailed results for ResNet18 UFL fine-tuned from ImageNet.} 
\label{tab:classificationDetail}
\begin{center}
\scalebox{0.55}{
\begin{tabular}{|l|l|l|l|l|} 
\hline
\rule[-1ex]{0pt}{3.5ex}  	\textbf{Class}	&	\textbf{Train Population}	&	\textbf{Test Population}	&	\textbf{Top-1}	&	\textbf{Top-5}	\\ 
\hline
\rule[-1ex]{0pt}{3.5ex}  	Building                	&	307221	&	100899	&	98.32	&	99.89	\\ 
\hline
\rule[-1ex]{0pt}{3.5ex}  	Building Aircraft Hangar 	&	180	&	97	&	4.12	&	64.95	\\ 
\hline
\rule[-1ex]{0pt}{3.5ex}  	Building Damaged        	&	1036	&	306	&	1.31	&	44.77	\\ 
\hline
\rule[-1ex]{0pt}{3.5ex}  	Building Facility       	&	823	&	316	&	2.85	&	71.2	\\ 
\hline
\rule[-1ex]{0pt}{3.5ex}  	Building Hut-Tent       	&	703	&	126	&	0	&	33.33	\\ 
\hline
\rule[-1ex]{0pt}{3.5ex}  	Building Shed           	&	1176	&	355	&	0.28	&	29.86	\\ 
\hline
\rule[-1ex]{0pt}{3.5ex}  	Construction Site        	&	1033	&	459	&	67.54	&	90.85	\\ 
\hline
\rule[-1ex]{0pt}{3.5ex}  	Container Lot            	&	2120	&	871	&	39.95	&	81.17	\\ 
\hline
\rule[-1ex]{0pt}{3.5ex}  	Engineering Vehicle (EV)     	&	204	&	46	&	4.35	&	21.74	\\ 
\hline
\rule[-1ex]{0pt}{3.5ex}  	EV Cementmixer          	&	287	&	87	&	2.3	&	49.43	\\ 
\hline
\rule[-1ex]{0pt}{3.5ex}  	EV Container Crane       	&	159	&	46	&	2.17	&	54.35	\\ 
\hline
\rule[-1ex]{0pt}{3.5ex}  	EV Crane                	&	173	&	46	&	4.35	&	10.87	\\ 
\hline
\rule[-1ex]{0pt}{3.5ex}  	EV Dump Truck            	&	1344	&	349	&	4.01	&	45.85	\\ 
\hline
\rule[-1ex]{0pt}{3.5ex}  	EV Excavator            	&	830	&	326	&	48.77	&	77.91	\\ 
\hline
\rule[-1ex]{0pt}{3.5ex}  	EV Grader               	&	83	&	39	&	2.56	&	35.9	\\ 
\hline
\rule[-1ex]{0pt}{3.5ex}  	EV Haul Truck            	&	325	&	40	&	35	&	95	\\ 
\hline
\rule[-1ex]{0pt}{3.5ex}  	EV Loader               	&	626	&	370	&	9.46	&	55.68	\\ 
\hline
\rule[-1ex]{0pt}{3.5ex}  	EV Mobile Crane          	&	313	&	76	&	0	&	26.32	\\ 
\hline
\rule[-1ex]{0pt}{3.5ex}  	EV Reach Stacker         	&	69	&	30	&	0	&	20	\\ 
\hline
\rule[-1ex]{0pt}{3.5ex}  	EV Straddle Carrier      	&	57	&	63	&	0	&	30.16	\\ 
\hline
\rule[-1ex]{0pt}{3.5ex}  	EV Tower Crane           	&	144	&	56	&	0	&	42.86	\\ 
\hline
\rule[-1ex]{0pt}{3.5ex}  	EV Tractor-Scraper      	&	78	&	19	&	0	&	5.26	\\ 
\hline
\rule[-1ex]{0pt}{3.5ex}  	Fixed Wing Aircraft (FWA)  	&	73	&	39	&	7.69	&	25.64	\\ 
\hline
\rule[-1ex]{0pt}{3.5ex}  	FWA Cargo                  	&	633	&	323	&	82.35	&	97.21	\\ 
\hline
\rule[-1ex]{0pt}{3.5ex}  	FWA Small               	&	354	&	110	&	48.18	&	84.55	\\ 
\hline
\rule[-1ex]{0pt}{3.5ex}  	Helicopter              	&	68	&	94	&	10.64	&	37.23	\\ 
\hline
\rule[-1ex]{0pt}{3.5ex}  	Helipad                 	&	120	&	55	&	38.18	&	69.09	\\ 
\hline
\rule[-1ex]{0pt}{3.5ex}  	Maritime Vessel (MV)        &	633	&	143	&	12.59	&	57.34	\\ 
\hline
\rule[-1ex]{0pt}{3.5ex}  	MV Barge                	&	171	&	62	&	0	&	35.48	\\ 
\hline
\rule[-1ex]{0pt}{3.5ex}  	MV Container            	&	271	&	78	&	38.46	&	92.31	\\ 
\hline
\rule[-1ex]{0pt}{3.5ex}  	MV Ferry                	&	183	&	16	&	12.5	&	56.25	\\ 
\hline
\rule[-1ex]{0pt}{3.5ex}  	MV Fishing              	&	723	&	289	&	13.15	&	50.52	\\ 
\hline
\rule[-1ex]{0pt}{3.5ex}  	MV Motor                	&	1447	&	154	&	22.08	&	59.09	\\ 
\hline
\rule[-1ex]{0pt}{3.5ex}  	MV Oil                  	&	64	&	23	&	8.7	&	65.22	\\ 
\hline
\rule[-1ex]{0pt}{3.5ex}  	MV Sail                 	&	692	&	24	&	20.83	&	50	\\ 
\hline
\rule[-1ex]{0pt}{3.5ex}  	MV Tug                  	&	209	&	45	&	8.89	&	75.56	\\ 
\hline
\rule[-1ex]{0pt}{3.5ex}  	MV Yacht                	&	430	&	7	&	0	&	42.86	\\ 
\hline
\rule[-1ex]{0pt}{3.5ex}  	Passenger Vehicle (PV)      &	2949	&	1299	&	1.08	&	48.42	\\ 
\hline
\rule[-1ex]{0pt}{3.5ex}  	PV Bus                  	&	6865	&	2047	&	25.6	&	81.63	\\ 
\hline
\rule[-1ex]{0pt}{3.5ex}  	PV SmallCar             	&	210827	&	61105	&	98.3	&	99.91	\\ 
\hline
\rule[-1ex]{0pt}{3.5ex}  	PV Pickup            	    &	1101	&	393	&	0	&	49.36	\\ 
\hline
\rule[-1ex]{0pt}{3.5ex}  	Pylon                   	&	349	&	124	&	41.13	&	80.65	\\ 
\hline
\rule[-1ex]{0pt}{3.5ex}  	Rail Vehicle (RV)          	&	17	&	2	&	0	&	0	\\ 
\hline
\rule[-1ex]{0pt}{3.5ex}  	RV Cargo                	&	1811	&	1843	&	64.35	&	91.05	\\ 
\hline
\rule[-1ex]{0pt}{3.5ex}  	RV Flat                 	&	123	&	70	&	28.57	&	67.14	\\ 
\hline
\rule[-1ex]{0pt}{3.5ex}  	RV Locomotive           	&	116	&	37	&	0	&	24.32	\\ 
\hline
\rule[-1ex]{0pt}{3.5ex}  	RV Passenger            	&	1567	&	383	&	27.68	&	81.2	\\ 
\hline
\rule[-1ex]{0pt}{3.5ex}  	RV Tank                 	&	120	&	83	&	36.14	&	80.72	\\ 
\hline
\rule[-1ex]{0pt}{3.5ex}  	Shipping Container       	&	1570	&	612	&	2.29	&	50	\\ 
\hline
\rule[-1ex]{0pt}{3.5ex}  	Storage Tank             	&	1625	&	807	&	40.89	&	82.78	\\ 
\hline
\rule[-1ex]{0pt}{3.5ex}  	Tower                    	&	84	&	19	&	0	&	10.53	\\ 
\hline
\rule[-1ex]{0pt}{3.5ex}  	Truck                   	&	12052	&	4369	&	7.87	&	78.51	\\ 
\hline
\rule[-1ex]{0pt}{3.5ex}  	Truck w/ Box Trailer      	&	3562	&	767	&	10.3	&	58.41	\\ 
\hline
\rule[-1ex]{0pt}{3.5ex}  	Truck Cargo             	&	5857	&	2035	&	2.01	&	57.05	\\ 
\hline
\rule[-1ex]{0pt}{3.5ex}  	Truck w/ Flatbed Trailer    &	883	&	262	&	1.15	&	23.66	\\ 
\hline
\rule[-1ex]{0pt}{3.5ex}  	Truck w/ Liquid Trailer    	&	145	&	55	&	0	&	7.27	\\ 
\hline
\rule[-1ex]{0pt}{3.5ex}  	Truck Tractor Trailer    	&	857	&	277	&	6.5	&	23.83	\\ 
\hline
\rule[-1ex]{0pt}{3.5ex}  	Truck Trailer           	&	4045	&	1254	&	4.31	&	48.64	\\ 
\hline
\rule[-1ex]{0pt}{3.5ex}  	Truck Utility           	&	3603	&	1705	&	0.35	&	49.85	\\ 
\hline
\rule[-1ex]{0pt}{3.5ex}  	Vehicle Lot              	&	3936	&	1124	&	46.71	&	91.28	\\
\hline
\end{tabular}}
\end{center}
\end{table}

\end{document}